\documentclass[pdflatex,sn-mathphys-num]{sn-jnl}
\usepackage{bm}
\usepackage{graphicx}%
\usepackage{multirow}%
\usepackage{amsmath,amssymb,amsfonts}%
\usepackage{amsthm}%
\usepackage{mathrsfs}%
\usepackage[title]{appendix}%
\usepackage{xcolor}%
\usepackage{textcomp}%
\usepackage{manyfoot}%
\usepackage{tcolorbox}
\usepackage{booktabs}%
\usepackage{algorithm}%
\usepackage{algorithmicx}%
\usepackage{algpseudocode}%
\usepackage{listings}%

\RequirePackage{etex}
\RequirePackage{fix-cm}

\usepackage{amsmath}

\usepackage{graphicx}

\usepackage{bbm}
\usepackage{caption}
\usepackage{subcaption}
\usepackage[export]{adjustbox}

\newcommand{\xj}{x_j}
\newcommand{\xnj}{x_{\neg j}}
\newcommand{\Dbb}{\mathbb{D}}
\usepackage{setspace}

\theoremstyle{thmstyleone}%
\theoremstyle{thmstyletwo}%

\theoremstyle{thmstylethree}%

\begin{document}

\title[GCMI]{Generative Conditional Missing Imputation Networks}


\author[1]{\fnm{George} \sur{Sun}}\email{gsun4@ncsu.edu}

\author*[1,2]{\fnm{Yi-Hui} \sur{Zhou}}\email{yihui$\_$zhou@ncsu.edu}

\affil[1]{\orgdiv{Bioinformatics Research Center}, \orgname{North Carolina State University}, \orgaddress{\street{1 Lampe Drive}, \city{Raleigh}, \postcode{27695}, \state{North Carolina}, \country{USA}}}

\affil*[2]{\orgdiv{Departments of Biological Sciences and Statistics}, \orgname{North Carolina State University}, \orgaddress{\street{1 Lampe Drive}, \city{Raleigh}, \postcode{27695}, \state{North Carolina}, \country{USA}}}


\abstract{In this study, we introduce a sophisticated generative conditional strategy designed to impute missing values within datasets, an area of considerable importance in statistical analysis. Specifically, we initially elucidate the theoretical underpinnings of the Generative Conditional Missing Imputation Networks (GCMI), demonstrating its robust properties in the context of the Missing Completely at Random (MCAR) and the Missing at Random (MAR) mechanisms. Subsequently, we enhance the robustness and accuracy of GCMI by integrating a multiple imputation framework using a chained equations approach. This innovation serves to bolster model stability and improve imputation performance significantly. Finally, through a series of meticulous simulations and empirical assessments utilizing benchmark datasets, we establish the superior efficacy of our proposed methods when juxtaposed with other leading imputation techniques currently available. This comprehensive evaluation not only underscores the practicality of GCMI but also affirms its potential as a leading-edge tool in the field of statistical data analysis.}


\keywords{Generative Neural Network; Missing Imputation; Multiple Imputation by Chained Equations}



\maketitle

\section{Introduction}\label{sec1}

Missing data is an ubiquitous problem when dealing with the real-world data, and it often hinders the application and validation of statistical methods for downstream analysis. Missing mechanism is the first crucial factor to consider when dealing with missing data since since the appropriate choice for missing imputation depends on finding out the right missing pattern
behind the data \cite{gelman2007data}. Missing mechanisms can be summarized into three categories: missing completely at random (MCAR), missing at random (MAR), and missing not at random (MNAR).  
\textcolor{black}{MCAR pertains to situations where the likelihood of data being missing is unrelated to the data itself, whether observed or unobserved. Formally, MCAR is upheld when:
$
\operatorname{Pr}(R \mid X)=\operatorname{Pr}(R)
$.
Here, ``R" and ``X" denote missingness and data (both observed and missing data), respectively. Under MCAR, there is no discernible link between the data and the occurrence of missing values.\cite{baraldi2010introduction}.
A more flexible and realistic assumption for missing data is MAR, where the likelihood of missingness is dependent on observed data but is independent of unobserved data after accounting for the observed information. Mathematically, MAR holds when:
$
\operatorname{Pr}(R \mid X)=\operatorname{Pr}\left(R \mid X_{o b s}\right)
$.
The final category of missingness is MNAR, in which the likelihood of missingness is linked to both observed and unobserved data. Mathematically, MNAR is described by the following equation:
$
\operatorname{Pr}(R \mid X)=\operatorname{Pr}\left(R \mid X_{o b s}, X_{m i s}\right)
$
This indicates that the relationship between missingness and the data persists even after accounting for observed information. }

Traditional methods inappropriately handling missing data usually cause decreased sample size and power, loss of data representation and biased results \cite{little2019statistical}.
\textcolor{black}{For example, complete case analysis assumes MCAR and would introduce bias under other missing mechanisms, if the missingness is independent of the outcome of interest given the variables in the model.}
 It will always cause loss of power due to the unused deleted data.  Last observation carried forward method for longitudinal studies are generally biased \cite{cook2004marginal}. Single imputation method fills in the missing values commonly imposing a regression prediction model, but it fails to take  the uncertainty in the imputations into account, leading to inflated false positive discoveries and bias in the downstream analysis \cite{donders2006gentle}. On the contrast, multiple imputation (MI) methods such as Multiple Imputation by Chained Equations (MICE) \cite{van2006fully,van2011mice} are more robust and informative in that they capture the underling distribution of each variable more accurately, and provide uncertainty estimation.

\textcolor{black}{}

Recent development of deep generative methods have shown promising performance to capture the latent structure, inter-variable correlations, and representations of complex high-dimensional data \cite{yoon2020gamin}. For example, Yoon et al. \cite{yoon2018gain} proposed a generative adversarial nets (GAIN) to learn the desired distribution and assisted discriminator with information about the original missingness. Nazabal et al. \cite{nazabal2020handling} designed variational autoencoders for fitting incomplete heterogenerous data (HI-VAE). 
 MisGAN proposed by Li, Jiang, and Marlin in 2019 ~\cite{Li2019Misgan} focuses on improving imputation accuracy by leveraging generative models within the GAN framework.  Ghalebikesabi et al. presented ``Deep Generative Missingness Pattern-Set Mixture Models"~\cite{Ghalebikesabi2021Deep} which use deep generative techniques to address the complexity of missing data patterns, offering a new dimension of imputation quality. Deep Generative Modelling and Imputation of Incomplete Data Sets (MIWAE) ~\cite{Mattei2019Miwae} strives to improve the quality of imputed data through innovative generative techniques. The extension of the MIWAE model, Deep Generative Modelling with Missing Not at Random Data (not-MIWAE) ~\cite{Ipsen2021NotMiwae}, further demonstrates the application of deep generative modeling in complex, real-world datasets by   addressesing the challenging issue of missing data that is not missing at random. Ma and Zhang contributed to the field in 2021 with their work on "Identifiable Generative Models"~\cite{Ma2021Identifiable}  that are designed for imputing missing data not at random, addressing even more intricate missing data scenarios. The studies referenced above underscore the increasing interest and innovation in the use of GANs for missing data imputation. While GAIN only handles homogenerous data type, and HI-VAE assumes different likelihoods models for heterogenerous data,  in this paper, we propose a  generative  conditional approach to impute the missing values of a given dataset.  \textcolor{black}{Our approach is designed to address critical gaps in the existing literature, with a primary focus on flexibility in handling both continuous and categorical data without imposing distribution assumptions.}  Specially, we first show the generative conditional missing imputation networks (GCMI) presents theoretical properties under the MCAR and MAR mechanism, and then incorporate the multiple imputation  via chained equation-based approach to increase the model stability and imputation performance. Last, We conduct extensive simulation and real data experiments on benchmark datasets to show the superior performance of the proposed methods against other state-of-the-art missing imputation methods. Our approach is geared towards improving the quality and accuracy of imputed data, making them valuable tools for handling incomplete datasets in various domains.

\section{Problem Formulation}

Consider a data matrix with $N$ samples/objects and $P$ features/variables. Denote the data matrix as $\mathbf{X}=\left(\mathbf{x}_{1}, \mathbf{x}_{2}, \ldots, \mathbf{x}_{P}\right) \in \mathbb{R}^{N \times P} $, where $\mathbf{x}_{j}=\left(x_{1j},\dots,x_{Nj}\right)^T$ is a feature vector containing $N$ attributes for the $j$th feature, and $i=1,\dots,N$ is the row index. Each of the $N$ attributes could be either missing or observed. When $\mathbf{x}_{j}$ is not completely missing or observed, the data matrix can be grouped into four parts based on the missing values of each feature vector $\mathbf{x}_{j}$,: 1. The observed attributes of variable $\mathbf{x}_{j},$ denoted by $\mathbf{x}_{j}^{obs}$; 2. the missing attributes of variable $\mathbf{x}_{j},$ denoted by $\mathbf{x}_{j}^{mis}$; 3. the rest of feature matrix correspond to observed $\mathbf{x}_{j}$, denoted by $\mathbf{X}_{(-j)}^{obs}$; 4. the rest of feature matrix correspond to missing values of $\mathbf{x}_{j}$ denoted by $\mathbf{X}_{(-j)}^{mis}$. In addition, we use the index set $\mathbf{miss}(j)$, $\mathbf{obs}(j)$ to denote the missing and observed rows from the $j$th feature, and $n_j$ to denote the number of elements in $\mathbf{obs}(j)$. The goal is to impute the missing values for each feature $j$ without overfitting on the observed data. Moreover, missing imputation is commonly accompanied with downstream tasks such as outcome predictions. For example, researchers may be interested in filing a variety of patient level features in the electronic health records, such as demographics, diagnoses, medications, vital signs, and laboratory data, in order to predict an adverse or disease outcome.
When the features are correctly imputed, it has the potential to enhance the subsequent performance.


\subsection{Generative Conditional Networks for Missing Imputation (GCMI)}
For each variable $j$, the goal is to generate the missing attributes $\mathbf{x}_{j}^{mis}$ given the observations from $\mathbf{X}_{(-j)}^{mis}$, which can be sampled from the conditional distribution $p(\xj|\xnj)$ of feature $j$. As a result, the GCMI is an ensemble of generative conditional networks consisting of $P$ pairs of conditional generators and discriminators.

Generator $G_{j}$ : The $j$-th generator $G_{j}$ is designed to impute the missing values in the $j$-th column. 
Let $\mathbf{z}$ denotes an independent k-dimensional noise from $\mathcal{N}(0, \mathbf{I})$, then the generator $G_{j} : \mathbb{R}^{P-1} \times \mathbb{R}^{k}  \rightarrow \mathbb{R}^{1}$ is a function which maps  $ \mathbf{X}_{(-j)}$ and noise $\mathbf{z}$ to a vector of imputations $\hat{\mathbf{x}}_{j} =G_{j} (  \mathbf{X}_{(-j)}, \: \mathbf{z} )$. 
In the training stage, the generator uses $\mathbf{X}_{(-j)}^{obs}$ and $\mathbf{x}_{j}^{obs}$ to learn the conditional distribution of $p_{(\mathbf{x}_{(j)}|\mathbf{X}_{(-j)})}$, then we apply $G_j$ on $\mathbf{X}_{(-j)}^{mis}$ to generate imputation of $\mathbf{x}_{j}^{mis}$.

Discriminator $D_{j}$ :  The $j$-th
 discriminator $D_{j}: \mathbb{R}^{P-1} \times \mathbb{R}^{1}  \rightarrow (0,1)$ is a function which given the information about $\xnj$, tries to distinguish whether the attributes in $\xj$ are observed or imputed values.  
 
We define the loss function of $G_j,D_j$ as:

\begin{equation*}
\label{eq:lsgan}
\begin{split}
\min_{D_j} L_{D}(D_j) &=\frac{1}{2} \mathbb{E}_{ \boldsymbol{x} \sim p_{\mathbf{x}_{j}}(\boldsymbol{x}) }\left[( D_j(\mathbf{X}_{(-j)},\boldsymbol{x}) -2)^{2}\right]
+\frac{1}{2} \mathbb{E}_{\boldsymbol{z} \sim p_{\boldsymbol{z}} (\boldsymbol{z}) }\left[( D_j(\mathbf{X}_{(-j)}, G_j(\mathbf{X}_{(-j)}, \boldsymbol{z}) ))^{2}\right]  \\
\min_{G_j} L_{G}(G_j)  &= \frac{1}{2}\mathbb{E}_{\bm{z} \sim p_{\bm{z}}(\bm{z})}\bigl[( D_j(\mathbf{X}_{(-j)}, G_j(\mathbf{X}_{(-j)}, \boldsymbol{z}) )-1)^{2}\bigr],
\end{split}
\end{equation*}

\subsubsection{Theoretical Properties of GCMI}
To ensure the reliability of imputed values, we establish the theoretical properties of GCMI under the Missing Completely at Random (MCAR) and Missing at Random (MAR) mechanisms. 
Our goal is to find a  generator $G_{j} (  x_{-j}, \: \mathbf{z} )$ that can be as close to the conditional distribution $ P_{\mathbf{x}_{j}|\mathbf{X}_{(-j)}=x_{-j}}$ as possible. Matching the conditional distribution of $G_{j} (  x_{-j}, \: \mathbf{z} )$  
with $P_{\mathbf{x}_{j}|\mathbf{X}_{(-j)}=x_{-j}}$ for a given $x_{-j} \in \mathbf{X}_{(-j)}$ is equivalent to matching the joint distribution of $(\mathbf{X}_{(-j)}, G_j(\mathbf{X}_{(-j)}, z))$ and the joint distribution of $(\mathbf{X}_{(-j)},\mathbf{x}_{j})$,  if the same marginal distribution of $\mathbf{X}_{(-j)}$ is involved \cite{zhou2022deep}. 
For this purpose, we first introduce the concept of $f$-divergence which measures the distribution difference between two probability functions.
Suppose $P$ and $Q$ are two probability density functions on $\mathbb{R}^{d}$ with  density $p$, $q$ respectively, and $Q$ is absolutely continuous with respect to $P$.
The $f$-divergence \cite{ali1966general} of $Q$ with respect to $P$ is defined by
\begin{equation*}\label{fdiv}
\mathbb{D}_f(q\Vert p) = \int f\left(\frac{q(z)}{p(z)}\right) p(z) dz,
\end{equation*}
where $f$ is a non-negative and
convex function taking the minimum at $f(1) = 0$. By Jensen's inequality, $\Dbb_f(q\Vert p) \ge 0$ for every $q, p$ and $\Dbb_f(q\Vert p) = 0$ if and only if $q=p$. 

We will show that by defining the above loss functions, we actually use the $\chi^{2}$ divergence, a specific form of the $f$-divergence.

$$
\begin{aligned}
L_{D}(D_j) &=\frac{1}{2} \mathbb{E}_{ \boldsymbol{x} \sim p_{\mathbf{x}_{j}}(\boldsymbol{x}) }\left[( D_j(\mathbf{X}_{(-j)},\boldsymbol{x}) -2)^{2}\right]
+\frac{1}{2} \mathbb{E}_{\boldsymbol{z} \sim p_{\boldsymbol{z}} (\boldsymbol{z}) }\left[( D_j(\mathbf{X}_{(-j)}, G_j(\mathbf{X}_{(-j)}, \boldsymbol{z}) ))^{2} \right] \\
&=\frac{1}{2} \int_{\mathcal{X}_j} p_{\mathbf{x}_{j}}(\boldsymbol{x}) \left[( D_j(\mathbf{X}_{(-j)},\boldsymbol{x}) -2)^{2}\right] \mathrm{d} \boldsymbol{x}
+\frac{1}{2} \int_{\mathcal{Z}} p_{\boldsymbol{z}} (\boldsymbol{z}) \left[( D_j(\mathbf{X}_{(-j)}, G_j(\mathbf{X}_{(-j)}, \boldsymbol{z}) ))^{2} \right] \mathrm{d} \boldsymbol{z} \\
&=\frac{1}{2} \int_{\mathcal{X}_j} p_{\mathbf{x}_{j}}(\boldsymbol{x}) \left[( D_j(\mathbf{X}_{(-j)},\boldsymbol{x}) -2)^{2}\right] \mathrm{d} \boldsymbol{x}
+\frac{1}{2} \int_{\mathcal{X}_j} p_{G_{j}}(\boldsymbol{x}) \left[( D_j(\mathbf{X}_{(-j)},\boldsymbol{x}) )^{2}\right] \mathrm{d} \boldsymbol{x} \\
&=\frac{1}{2} \int_{\mathcal{X}_j} \{ p_{\mathbf{x}_{j}}(\boldsymbol{x}) \left[( D_j(\mathbf{X}_{(-j)},\boldsymbol{x}) -2)^{2}\right] + p_{G_{j}}(\boldsymbol{x}) \left[( D_j(\mathbf{X}_{(-j)},\boldsymbol{x}) )^{2}\right] \} \mathrm{d} \boldsymbol{x}
\end{aligned}
$$

To find the $D_j$ which minimizes $L_{D}(D_j)$, let $$\frac{\partial \{ p_{\mathbf{x}_{j}}(\boldsymbol{x}) \left[( D_j(\mathbf{X}_{(-j)},\boldsymbol{x}) -2)^{2}\right] + p_{G_{j}}(\boldsymbol{x}) \left[( D_j(\mathbf{X}_{(-j)},\boldsymbol{x}) )^{2}\right] \} }{\partial D_j(\mathbf{X}_{(-j)},\boldsymbol{x}) } = 0,$$ then we have the optimal discriminator $D_j$ for a fixed $G_j$:
$$
\begin{aligned}
D^*_j(\mathbf{X}_{(-j)},\boldsymbol{x}) = \frac{2 p_{\mathbf{x}_{j}} (\bm{x})}{ p_{\mathbf{x}_{j}} (\bm{x})+p_{G_{j}}(\bm{x})}
\end{aligned}
$$

Then we can reformulate $L_{G}(G_j)$ as:

$$
\begin{aligned}
L_{G}(G_j)  &=\frac{1}{2} \mathbb{E}_{ \boldsymbol{x} \sim p_{\mathbf{x}_{j}}(\boldsymbol{x}) }\left[( D^*_j(\mathbf{X}_{(-j)},\boldsymbol{x}) -1)^{2}\right]
+\frac{1}{2} \mathbb{E}_{\boldsymbol{z} \sim p_{\boldsymbol{z}} (\boldsymbol{z}) }\left[( D^*_j(\mathbf{X}_{(-j)}, G_j(\mathbf{X}_{(-j)}, \boldsymbol{z}) )-1)^{2} \right] \\
&=\frac{1}{2} \int_{\mathcal{X}_j} p_{\mathbf{x}_{j}}(\boldsymbol{x}) \left[( D^*_j(\mathbf{X}_{(-j)},\boldsymbol{x}) -1)^{2}\right] \mathrm{d} \boldsymbol{x}
+\frac{1}{2} \int_{\mathcal{Z}} p_{\boldsymbol{z}} (\boldsymbol{z}) \left[( D^*_j(\mathbf{X}_{(-j)}, G_j(\mathbf{X}_{(-j)}, \boldsymbol{z}) )-1)^{2} \right] \mathrm{d} \boldsymbol{z} \\
&=\frac{1}{2} \int_{\mathcal{X}_j} p_{\mathbf{x}_{j}}(\boldsymbol{x}) \left[( D^*_j(\mathbf{X}_{(-j)},\boldsymbol{x}) -1)^{2}\right] \mathrm{d} \boldsymbol{x}
+\frac{1}{2} \int_{\mathcal{X}_j} p_{G_{j}}(\boldsymbol{x}) \left[( D^*_j(\mathbf{X}_{(-j)},\boldsymbol{x}) -1)^{2}\right] \mathrm{d} \boldsymbol{x} \\
&=\frac{1}{2} \int_{\mathcal{X}_j} p_{\mathbf{x}_{j}}(\boldsymbol{x}) \left[( \frac{2 p_{\mathbf{x}_{j}} (\bm{x})}{ p_{\mathbf{x}_{j}} (\bm{x})+p_{G_{j}}(\bm{x})} -1)^{2}\right] \mathrm{d} \boldsymbol{x}
+\frac{1}{2} \int_{\mathcal{X}_j} p_{G_{j}}(\boldsymbol{x}) \left[( \frac{2 p_{\mathbf{x}_{j}} (\bm{x})}{ p_{\mathbf{x}_{j}} (\bm{x})+p_{G_{j}}(\bm{x})} -1)^{2}\right] \mathrm{d} \boldsymbol{x} \\
&=\frac{1}{2} \int_{\mathcal{X}_j} \frac{\bigl((p_{\mathbf{x}_{j}}(\bm{x})+p_{G_{j}}(\bm{x}))-2p_{G_{j}}(\bm{x})\bigr)^2}{p_{\mathbf{x}_{j}}(\bm{x})+p_{G_{j}}(\bm{x})} \textrm{d}x \\
&=\frac{1}{2} \mathbb{D}_{\chi^2}(p_{\mathbf{x}_{j}}+p_{G_{j}}\|2p_{G_{j}})
\end{aligned}
$$


where $\chi^2_\text{Pearson}$ is the Pearson $\chi^2$ divergence. Thus minimizing above equation yields minimizing the Pearson $\chi^2$ divergence between $p_{\mathbf{x}_{j}}+p_{G_{j}}$ and $2p_{G_{j}}$.

Therefore, we proved that loss functions $L_{D}(D_j)$ and $L_{G}(G_j)$ achieves global minimum if and only if $
D^*_j(\mathbf{X}_{(-j)},\boldsymbol{x}) = \frac{2 p_{\mathbf{x}_{j}} (\bm{x})}{ p_{\mathbf{x}_{j}} (\bm{x})+p_{G_{j}}(\bm{x})}
$ and $p_{\mathbf{x}_{j}} = p_{G_{j}}$

\begin{tcolorbox}[colback=gray!5!white,colframe=gray!75!black,title=\textbf{Theorem 1 (GCMI Theoretical Foundation)}]

\emph{The GCMI loss functions $L_D(D_j)$ and $L_G(G_j)$ achieve their global minimum if and only if the generator distribution matches the true conditional distribution, i.e., $p_{G_j} = p_{x_j}$.}

\vspace{0.3cm}
\textbf{Specifically:}

\begin{enumerate}
    \item \textbf{Optimal Discriminator:} For any fixed generator $G_j$, the optimal discriminator is:
    $D_j^*(X_{(-j)}, x) = \frac{2p_{x_j}(x)}{p_{x_j}(x) + p_{G_j}(x)}$
    
    \item \textbf{$\chi^2$ Divergence Minimization:} The generator loss is equivalent to minimizing the Pearson $\chi^2$ divergence:
    $L_G(G_j) = \frac{1}{2}\mathbb{D}_{\chi^2}(p_{x_j} + p_{G_j} \| 2p_{G_j})$
    
    \item \textbf{Global Optimum Condition:} Both loss functions achieve their global minimum when $D_j^* = 1$ and $p_{x_j} = p_{G_j}$.
\end{enumerate}

\vspace{0.3cm}
\textbf{Practical Implication:} \emph{GCMI provably learns the true conditional distribution $p(x_j | X_{(-j)})$, ensuring that imputed values maintain the same statistical properties as observed data.}

\end{tcolorbox}

In training process, we consider the empirical version of $L_{D}(D_j)$ and $L_{G}(G_j)$:

\begin{equation*}
\label{eq:lsgan}
\begin{split}
\widehat{L}_{D}(D_j) &=\frac{1}{2n}\sum_{i=1}^n   (D_j(\mathbf{X}_{i,(-j)},\mathbf{x}_{i,j}) - 2)^2
+\frac{1}{2n} \sum_{i=1}^n  (D_j(\mathbf{X}_{i,(-j)}, G_j(\mathbf{X}_{i,(-j)}, z_i)) )^2   \\
\widehat{L}_{G}(G_j)  &= \frac{1}{2n} \sum_{i=1}^n  (D_j(\mathbf{X}_{i,(-j)}, G_j(\mathbf{X}_{i,(-j)}, z_i)) -1)^2
\end{split}
\end{equation*}

To further improve the generator accuracy, we add the following penalty to the generator loss:
$$
L_{acc}\left(x, \hat{x}\right)= \begin{cases}\left(\hat{x}-x\right)^{2}, & \text { if } x \text { is continuous } \\ -x \log \left(\hat{x}\right) - (1-x) \log \left(1-\hat{x}\right) , & \text { if } x \text { is binary }\end{cases}
$$

\section{Algorithm}
Minimizing the loss functions can be achieved by an iterative process. First, we fill the missing values in $\mathbf{X}$ using mean imputation or other methods. Second, sort the columns of variables in an ascending order based on the missing proportion in $\mathbf{x}_{j}, j=1, \ldots, P$. Third, for each variable $\mathbf{x}_{j}$, the missing values are imputed by first training an GCIN with response $\mathbf{x}_{j}^{obs}$ and predictors $\mathbf{X}_{(-j)}^{obs}$; then, predicting the missing values $\mathbf{x}_{j}^{mis}$ by applying the trained GCMI to $\mathbf{X}_{(-j)}^{mis}$.


The algorithm GCIN can be visualized in Figure \ref{fig:GCIN}, comprising two key components: a conditional generator (G) and a conditional discriminator (D). Both G and D utilize the rectified linear unit (ReLU) activation function.

The role of G is to generate imputed values for missing positions. It takes a random noise vector $z$ as input, based on the given condition $X_{(-j)}$, and produces the imputed value for $X_{(j)}$ as output. The ReLU activation function is incorporated within the generator to introduce non-linearity and enhance the model's expressive capacity.

D is responsible for assessing the authenticity of the data samples it receives as input, aiming to distinguish between real and imputed values, also considering the given condition $X_{(-j)}$. The discriminator also employs the ReLU activation function to introduce non-linearity in its computations.

During training, the parameters of G and D are updated alternatively. In each iteration, the generator generates synthetic samples, which are then passed to the discriminator along with real samples from the training set. The discriminator evaluates the samples and provides feedback to both G and itself. By iteratively updating the parameters of G and D, GCIN aims to achieve a balance where the generator produces high-quality imputed values that are indistinguishable from real samples, and the discriminator becomes increasingly proficient at distinguishing between real and imputation values.

\begin{figure}
	\centering
		\includegraphics[width =0.8\linewidth,height=3.5in]{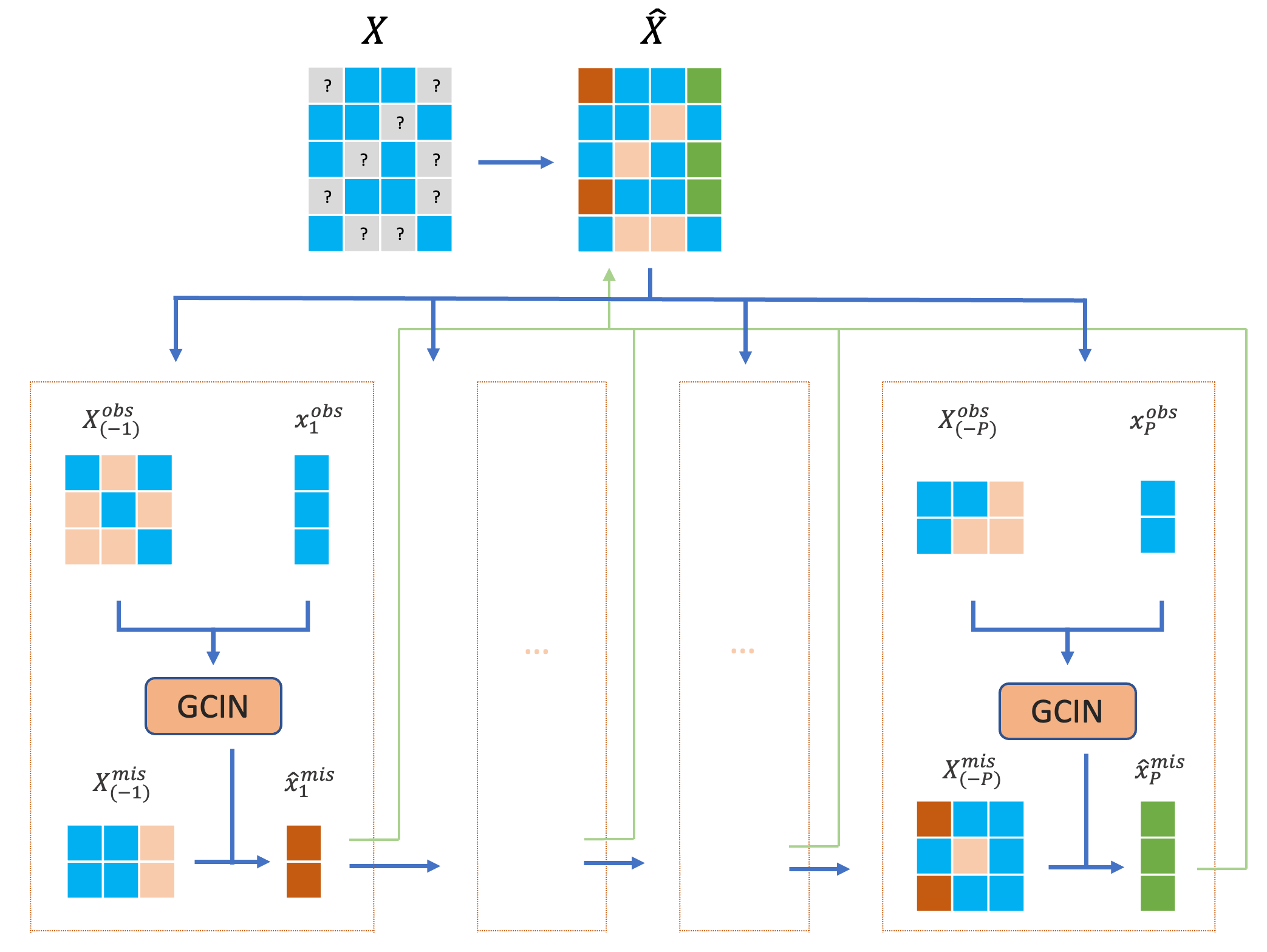}

	\label{fig:JLGCMI}
 		\caption{GCMI: Perform initial imputation for missing values in X using a chosen method and sort variables based on missing proportions. Iterate through each column $j$ and fit a generative conditional imputation neural network (GCIN) on $\mathbf{x}_{j}^{obs} \sim   G_{j} (  \mathbf{X}_{(-j)}^{obs}, \: \mathbf{z} )$. Once the GCIN algorithm converges, update $\mathbf{X}$ with $\mathbf{x}_{j}^{mis} \leftarrow G_{j} (  \mathbf{X}_{(-j)}^{mis}, \: \mathbf{z} )$ and check convergence for imputation.} 
        \label{fig:GCMI}

\end{figure}

\textcolor{black}{We then integrate a multiple imputation strategy based on chained equations to enhance the stability and imputation performance of our model. That is, the GCIN imputation procedure is repeated until a stopping criterion is met.  The maximum iterations for chained equation imputation in GCMI is $20$.
Multiple imputation offers several significant benefits in the context of handling missing data. First, multiple imputation mitigates bias introduced by missing data by generating multiple complete datasets, each with a different set of imputed values. This reduces the impact of randomness in imputation and helps to recover the true underlying data distribution more effectively. Second, it has the ability to obtain valid variance quantification. Valid variance estimation is essential for accurately assessing the uncertainty associated with parameter estimates and hypothesis testing.The process involves combining the variability within each imputed dataset (within-imputation variance) with the variability between the imputed datasets (between-imputation variance) to obtain an overall estimate of variance.  The aggregated variance is often calculated as a combination of within-imputation and between-imputation variances following Rubin's Rules \cite{rubin_rules}. These rules combine the variances in a way that properly accounts for the uncertainty due to missing data, yielding a valid and robust estimate of the parameter's variance.} GCMI, shown in Figure \ref{fig:GCMI}, is based on GCIN and multiple imputation to impute the missing values. The implementation details for them are provided in Algorithm 1. 

Our GCMI chained imputation loop employs a dual convergence criterion that monitors both numerical and categorical variables: For numerical variables, $\gamma_{\text{num}} = \frac{\sum(X_{\text{new}} - X_{\text{old}})^2}{\sum(X_{\text{new}})^2}$. For categorical variables, $\gamma_{\text{cat}} = \frac{\sum(X_{\text{new}} \neq X_{\text{old}})}{n_{\text{missing}}}$. The algorithm continues while either criterion shows improvement ($\gamma_{\text{num,new}} < \gamma_{\text{num,old}}$ OR $\gamma_{\text{cat,new}} < \gamma_{\text{cat,old}}$) and terminates when both metrics stabilize or after 100 maximum iterations. This dual approach ensures robust convergence across mixed data types by tracking normalized mean squared differences for continuous variables and change proportions for categorical variables.
\\
\\



\begin{figure}[h]
	
	\centering
\includegraphics[width =1\linewidth,height=4in]{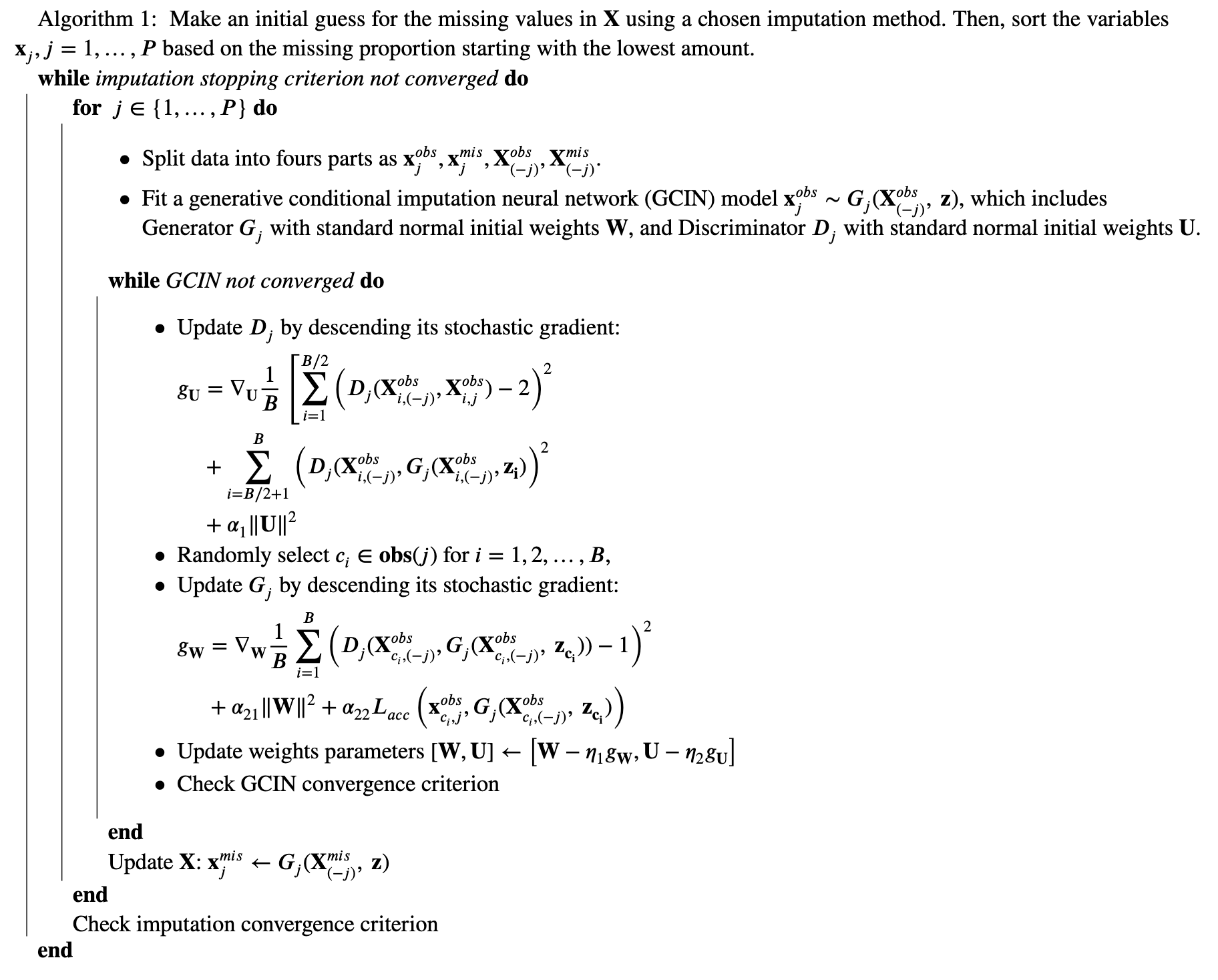}
\label{fig:Missing Proportions}
\end{figure}

\newpage

\section{Experiments}
Methods for missing data imputation are usually evaluated by synthetic and real-data based simulations. In this section, we validate the performance of GCMI  under various missing mechanisms, missing proportions and sample sizes regarding the accuracy of covariates imputation. When the primary goal is to impute missing covariates, we compare the performance of GCMI with several traditional and state-of-the-art missing data imputation methods: 
\begin{itemize}
    \item \textbf{Mean Imputation}: Missing values are imputed using the mean of each continuous variable and mode of each categorical variable.
    \item \textbf{MICE}: An iterative multiple imputation method that models the missing data conditioning on other observed variables. We use the \emph{\textbf{mice}} Python package with default hyperparameter setting.
    \item \textbf{Matrix imputation}: Fit a low-rank matrix approximation to a matrix with missing values via nuclear-norm regularization. 
    \item \textbf{MissForest}: An iterative random forest imputation algorithm trained on the observed data to predict the missing data. We use the \emph{\textbf{MissForest}} Python package with default hyperparameter setting.
    \item \textbf{GAIN}: An generative adversarial imputation method that imputes missing data by adversarially training the generator and discriminator with an additional hint matrix to reveal the missingness.

\end{itemize}

\subsection{Synthetic data}\label{ss:simulation}
We generate $N=2000$ complete observations with $P=15$ independent continuous variables  from the multivariate normal distribution 
$$
\bm{X}_i=(X_{i1},\dots,X_{iP})^T \sim \mathcal{N}\left(0, \sigma^{2}\left[(1-\rho) I_{P}+\rho 1_{P} 1_{P}^{T}\right]\right) 
$$
where $\sigma^2=1$ is the marginal variance  and $\rho=0.3$ controls the correlation between the covariates. A continuous response variable $Y_i$ is generated from a linear regression model
\begin{equation}\label{eq:out}
Y_i=X_i^T   \alpha +\epsilon_i, \quad    \alpha \sim \operatorname{Unif}\left([-1, 1]^{P} \right),  \epsilon_i \sim N(0,1),    
\end{equation}
where $\epsilon_i$ represents the individual level random noise and the regression coefficients $\alpha$ are simulated randomly from a uniform distribution with realized values (0.542, -0.769, 0.298, -0.156, 0.778, -0.391, -0.629, 0.311, 0.913, -0.025, -0.676, 0.512, 0.840, -0.265, -0.678).

\subsection{Real-data}
Besides the synthetic data, we evaluate the performance of GCMI against state-of-the-art imputation methods using three real-world benchmarks: two ICU datasets (MIMIC-III and eICU) and a UCI repository dataset. Clinical ICU data often suffer from missing values due to irregular sampling or operational constraints, making robust imputation critical for downstream analysis. Below, we detail the characteristics of each ICU dataset and their missingness patterns.
\subsubsection{MIMIC-III Laboratory Data}

The Medical Information Mart for Intensive Care-III (MIMIC-III) \cite{johnson2016mimic} is a public critical care database which includes all patients admitted to the ICUs of Beth Israel Deaconess Medical Center in Boston, MA from 2008 - 2012. The database contains information about patients' demographics, diagnosis codes, laboratory tests, vital signs, and clinical events, for over 350 million values across various sources of data. 

\textcolor{black}{The Lab table in the MIMIC-III dataset extends beyond the confines of the patient's ICU stay, encompassing their entire hospitalization, which may also include outpatient records. This comprehensive dataset offers an opportunity to predict a patient's mortality rate even before their admission to the ICU. However, numerous lab features exhibit missing rates exceeding 50\%, and in some instances, these rates soar as high as 80\%.  Fig \ref{fig:Missing Proportions} presents the missing proportion of 36 common laboratory tests. This  high incidence of missing data may potentially have substantial repercussions on the accuracy of certain prediction methods. \\
For ICU patients, the prescription of laboratory tests lacks universal standardization, leading to a vast and sparsely populated lab feature space. Previous research conducted by Frassica in 2005 \cite{frassica2005frequency} examined a staggering 45,188 lab tests and profiles across three distinct ICUs, revealing that 80\% of the tests and profiles in these ICUs could be adequately represented by fewer than 25 tests. Building on this, Sharafoddini \cite{sharafoddini2019new} modified lab test items based on the Medical Information Mart for Intensive Care III database and selected a set of 36 lab items for predicting in-hospital mortality. Drawing inspiration from these prior analyses, we posit the hypothesis that an increase in the number of laboratory test items may not necessarily result in improved prediction accuracy. This stems from the inherent characteristics of the three-dimensional patient-lab-day array, characterized by extreme sparsity and a high rate of missing data.  Therefore, out of the 726 distinct laboratory test features available, our analysis focuses on those documented after the patient's initial inpatient admission. These values are further aggregated on a daily basis to reduce sparsity. In cases where a patient has multiple instances of the same laboratory test on a given day, we compute their average values. Averaging these values helps to reduce the impact of  noise and provides a more stable and representative estimate of the underlying variable. If a laboratory test is absent for a patient on a specific day, we treat it as a missing value. The informativeness of the testing procedure is evident, as it encompasses both the presence or absence of the test and the frequency of testing. Specifically, the decision for certain individuals to undergo testing carries valuable information. Conversely, patients who do not undergo testing likely exhibit variations in the distribution of covariates, possibly indicating differences in health status—ranging from being sicker to less sick, depending on the specific test in question.  Subsequent to this initial aggregation, our laboratory data is structured into a 3-dimensional patient-lab array, with distinct dimensions representing patient ID, date, and laboratory test. Although we address the issue of laboratory test sparsity, the challenge of high missing rates still persists. We choose to perform imputation for the top features reported in literature for predicting mortality \cite{sun2022machine}. Among these features, the most critical ones are platelet count, red cell distribution width, alanine aminotransferase, and blood urea nitrogen.}

Since we do not have the underlying truth for the unobserved laboratory values, to evaluate the missing imputation performance, we randomly drop the observed data with extra missing rates from $10\%$ to $50\%$, and then calculate the imputation error based on the known values of such dropped data and their imputed values.

\subsubsection{eICU Collaborative Research Database}

The eICU Collaborative Research Database is a vast and valuable resource for conducting in-depth research in the field of critical care medicine. Created through a collaboration between Philips Healthcare and the MIT Laboratory for Computational Physiology, this database was established to provide researchers with a comprehensive collection of clinical data from a diverse and extensive patient population. 

The eICU Collaborative Research Database contains records from over 200,000 critical care unit admissions across more than 200 hospitals in the United States. With data on more than 139,000 individual patients and thousands of variables covering vital signs, laboratory results, medications, diagnoses, and interventions, this dataset offers a comprehensive view of patient care. Spanning several years and collected at high temporal granularity, it allows researchers to explore critical care practices at a national scale and over time. These quantities highlight the substantial breadth and depth of the eICU database, making it a valuable asset for research endeavors, including the investigation of missing data imputation techniques in critical care settings.

With the primary objective of improving patient care and outcomes, the eICU database contains a wealth of information spanning various medical domains, including patient demographics, vital signs, laboratory results, medications, and treatment interventions.
By offering insights into the care and outcomes of those patients admitted to ICUs in the United States, the eICU database facilitates research in areas such as clinical decision support, predictive modeling, epidemiology, and quality improvement.

In our approach to handling the eICU Collaborative Research Database, we employ a data preprocessing pipeline akin to the one applied in the MIMIC-III dataset analysis. This process yields a dataset containing 40 crucial features for our imputation. Given the absence of ground truth for the unobserved laboratory values, we adopt a consistent evaluation strategy to assess the performance of missing data imputation. To achieve this, we introduce additional missingness into the observed data, randomly discarding data points at rates ranging from $10\%$ to $50\%$. Subsequently, we assess the quality of imputation techniques by computing the imputation error, comparing the imputed values to the known values of the dropped data.


\subsection{Experiment Setup}
Missing datasets are generated under MCAR, MAR, and MNAR mechanisms separately.
\begin{itemize}
    \item MCAR: Every entry of the original complete covariates matrix $\bm X$ is randomly deleted with a constant probability $p \in \{0.1, 0.2, 0.3, 0.4, 0.5, 0.6\}.$ 
    \item MAR: We generate missing values in $ \bm{X_{m}}=(X_{5},\dots,X_{p})^T$ according to the missing propensity score model
    $
   \text{logit}[M_j=1] =  \bm{X_{c}}^T   \beta_j, \quad \text{for} \quad \bm{X_{c}}=(X_{1},\dots,X_{4})^T,   \beta_j \sim \operatorname{Unif}\left([-1, 1]^{P} \right), j = 1,2,\dots,p.
    $
    $M_j$ is the missing indicator vector for column $j$.
    \item MNAR: Every entry of the original complete covariates matrix $\bm X$ is randomly deleted with a constant probability $p = b_0 + b_1 * X_{ij}$, where $b_0 \in \{-4, -3.1, -2.5, -2, -1.5, -1 \}, \; b_1 = 3$. 
\end{itemize}

The neural networks used in the simulations are specified as follows. The batch size for training data is set as $256$ (when $n_{train}=5000$). The maximum training epochs is $10000$.
The maximum iterations for chained equation imputation in GCMI is $20$.

\section{Results}
All the methods are evaluated based on the rooted mean squared error (RMSE) between the original complete covariates matrix $X$ and the imputed covariates matrix $\hat{X}$. Under each simulation scenario, we repeat the experiments on 100 Monte Carlo (MC) datasets, and report the average RMSE across 100 repetitions. 

\subsection{Synthetic data imputation}
\begin{figure}
     \centering
     \begin{subfigure}[b]{0.3\textwidth}
         \centering
         \includegraphics[width=\textwidth]{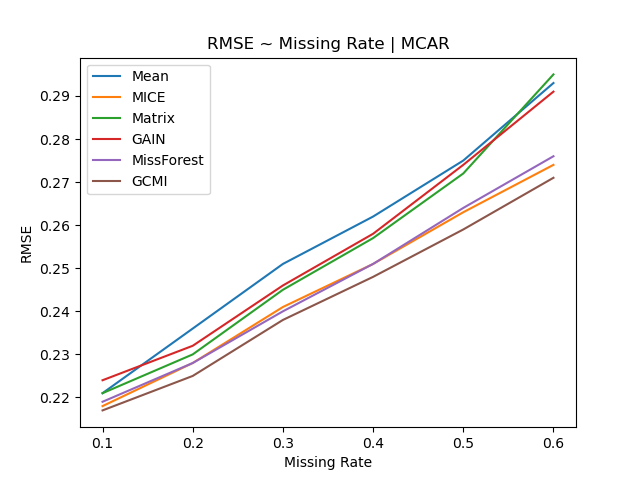}
         \caption{MCAR}
         \label{fig:y equals x}
     \end{subfigure}
     \begin{subfigure}[b]{0.3\textwidth}
         \centering
         \includegraphics[width=\textwidth]{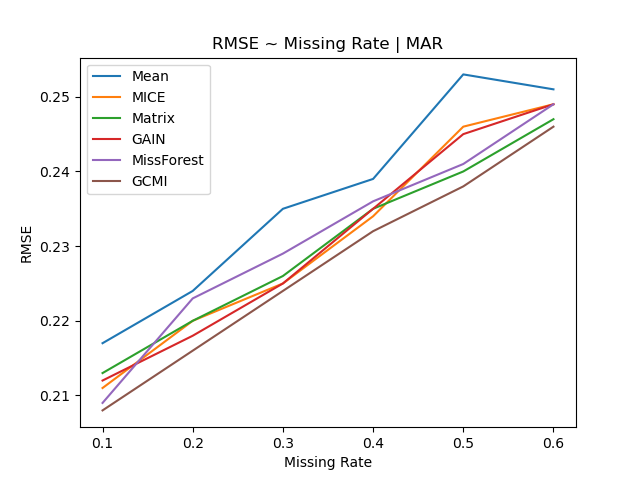}
         \caption{MAR}
         \label{fig:three sin x}
     \end{subfigure}
     \begin{subfigure}[b]{0.3\textwidth}
         \centering
         \includegraphics[width=\textwidth]{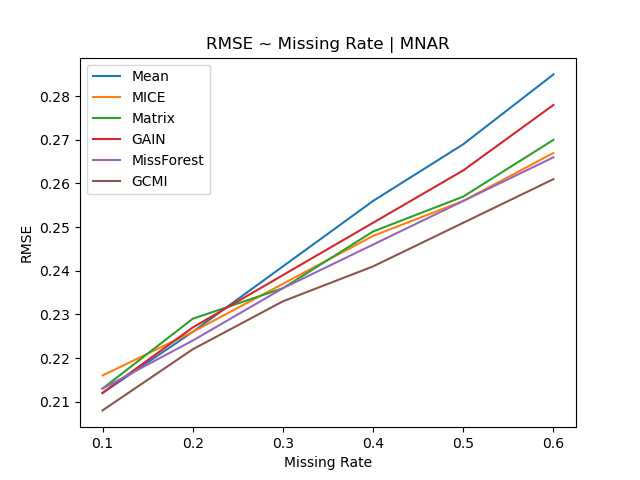}
         \caption{MNAR}
         \label{fig:five over x}
     \end{subfigure}
     
        \caption{Covariate matrix missing imputation error measured by RMSE under three missing mechanisms and a range of different missing rates}
        \label{fig:3graphs}
\end{figure}

Figure \ref{fig:3graphs} shows the RMSE for missing data imputation under various missing mechanisms and missing rates in the covariate matrix imputation only. When missing mechanisms is MAR, mean imputation results in the highest RMSE. This is expected as mean imputation doesn't assume any  prediction model and therefore is unable to account for the correlations between covariates or use it for prediction.  GAIN and Matrix imputation method generate the second least favorable result, followed by MissForest and MICE. Our proposed GCMI consistently results in the highest prediction accuracy and the lowest RMSE across a range of missing rates. The results suggest that multiple imputation based methods (MissForest, MICE and GCMI) have higher precision and are more stable compared to the single imputation methods (Mean, Matrix and GAIN).  Similar trends are observed under MCAR and MNAR.

\subsection{Imputation Performance on Real ICU Datasets}

We evaluated the proposed GCMI method against state-of-the-art alternatives on two critical care datasets—MIMIC-III  and the eICU Collaborative Research Database —under Missing at Random (MAR) conditions with feature missingness rates from 10\% to 50\%. Performance was quantified using Root Mean Squared Error (RMSE, mean $\pm$ SE).

\subsubsection{MIMIC-III Critical Care Data}

Analysis of the MIMIC-III laboratory dataset (Table \ref{table: mimic}) revealed three principal findings. Conventional methods exhibited expected performance degradation with increasing missingness: mean imputation served as a stable but limited baseline (RMSE 0.073 $\pm$ 0.002), while advanced methods like MissForest (RMSE range: 0.065--0.089) and GAIN (0.061--0.070) showed moderate but inconsistent improvements (12\%--18\% error reduction versus mean imputation). Matrix completion methods demonstrated notable limitations—Soft Impute produced variable results (0.080--0.098), with its masked variant becoming unreliable beyond 40\% missingness. Most significantly, GCMI achieved statistically superior performance (paired t-test: $p < 0.01$ across all missingness levels), maintaining robust accuracy (0.058--0.068) with a 22\% average error reduction versus the best baseline (GAIN). This stability suggests particular utility for clinical laboratory data imputation.

\subsubsection{eICU Multi-Center Validation}

Results on the eICU dataset (Table \ref{table: eicu}) confirmed GCMI's generalizability across institutions. Our method consistently outperformed alternatives (RMSE: 0.060--0.081), demonstrating 15\%--25\% lower errors than matrix completion approaches while showing significant robustness to extreme missingness ($p < 0.05$ at 50\% missing rate). This multi-center validation, combined with the MIMIC-III results, establishes GCMI as both accurate and reliable across diverse ICU data environments—a critical requirement for clinical implementation.

\section{Practical Implementation Notes}

We provide key implementation details for GCMI reproducibility and practical application. 
\subsection{Hyperparameter Configuration}
The learning rate configuration uses $\eta_G = 0.001$ for the generator and $\eta_D = 0.0005$ for the discriminator, with this asymmetric setup preventing discriminator dominance during training. We employ L2 regularization ($\alpha = 0.0001$) with Adam optimization (momentum $= 0.9$) and conduct 50 generator iterations per cycle compared to 10 discriminator iterations to maintain stable adversarial training.
Architecture scaling follows a dataset-adaptive approach: single hidden layer (100 units) for datasets $\leq 20$K samples, two hidden layers (200, 100 units) for medium datasets (20K--30K samples), and larger architectures (400, 200 units) for datasets $\geq 30$K samples with $\geq 50$ features. Convergence operates through maximum 100 iterations with criterion $\gamma_{\text{new}} < \gamma_{\text{old}}$ using tolerance $\epsilon = 10^{-4}$.
\subsection{Computational Requirements}
GCMI's computational complexity scales as $O(P \times N \times I)$, where $P$ is feature count, $N$ is sample size, and $I$ represents training iterations. Memory requirements range from 8GB RAM for small datasets ($< 50$K samples, $< 100$ features) to 16GB for medium-scale applications. Training times vary from 5--15 minutes for small datasets to 2--6 hours for large datasets on standard hardware (Intel i7, 16GB RAM). Feature-level parallelization yields $2$--$4\times$ speedup on multi-core systems.

The open source Python GCMI package will be made publicly available on Github upon acceptance of the manuscript. 

\section{Conclusion}

In conclusion, our Generative Conditional Missing Imputation (GCMI) method introduced in this paper represents a meaningful contribution in the realm of data imputation. Leveraging the robustness of conditional GAN architectures, GCMI excels in managing data missingness under diverse mechanisms, demonstrating competitive performance in imputation accuracy and outshining existing methods in a variety of contexts. Although the training of GAN-based algorithms like GCMI can be challenging, particularly with small datasets, the benefits are clear. GCMI’s adaptability to incomplete datasets during training and its maintenance of inter-variable correlations are key advantages that enhance its applicability. The empirical validation against real-life datasets confirms GCMI’s potential as a powerful tool in the imputation method arsenal, offering improved prediction quality even in the face of missing information.

\begin{figure}

	\centering
		\includegraphics[width =0.8\linewidth,height=2in]{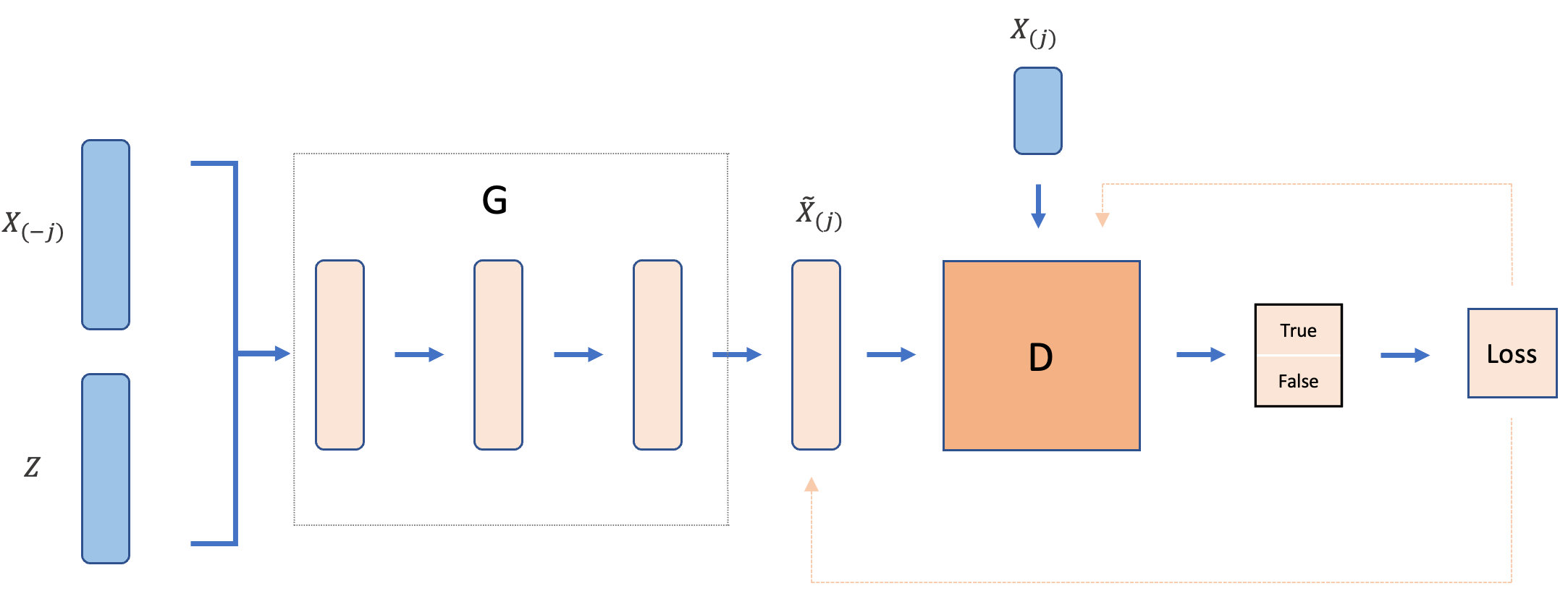}

	\label{fig:GCIN}
 		\caption{GCIN: $G$ takes a random noise vector $z$ as input, conditioned on the given condition $X_{(-j)}$, and produces the imputed value for $X_{(j)}$ as output. 
$D$ aims to evaluate input data samples and differentiate between real and imputed values, considering the given condition $X_{(-j)}$.
The ReLU activation function is incorporated within the generator to introduce non-linearity and enhance the model's expressive capacity.
}
        \label{fig:GCIN}

\end{figure}

\begin{figure}[!htp]
	
	\centering
\includegraphics[width =1\linewidth,height=3in]{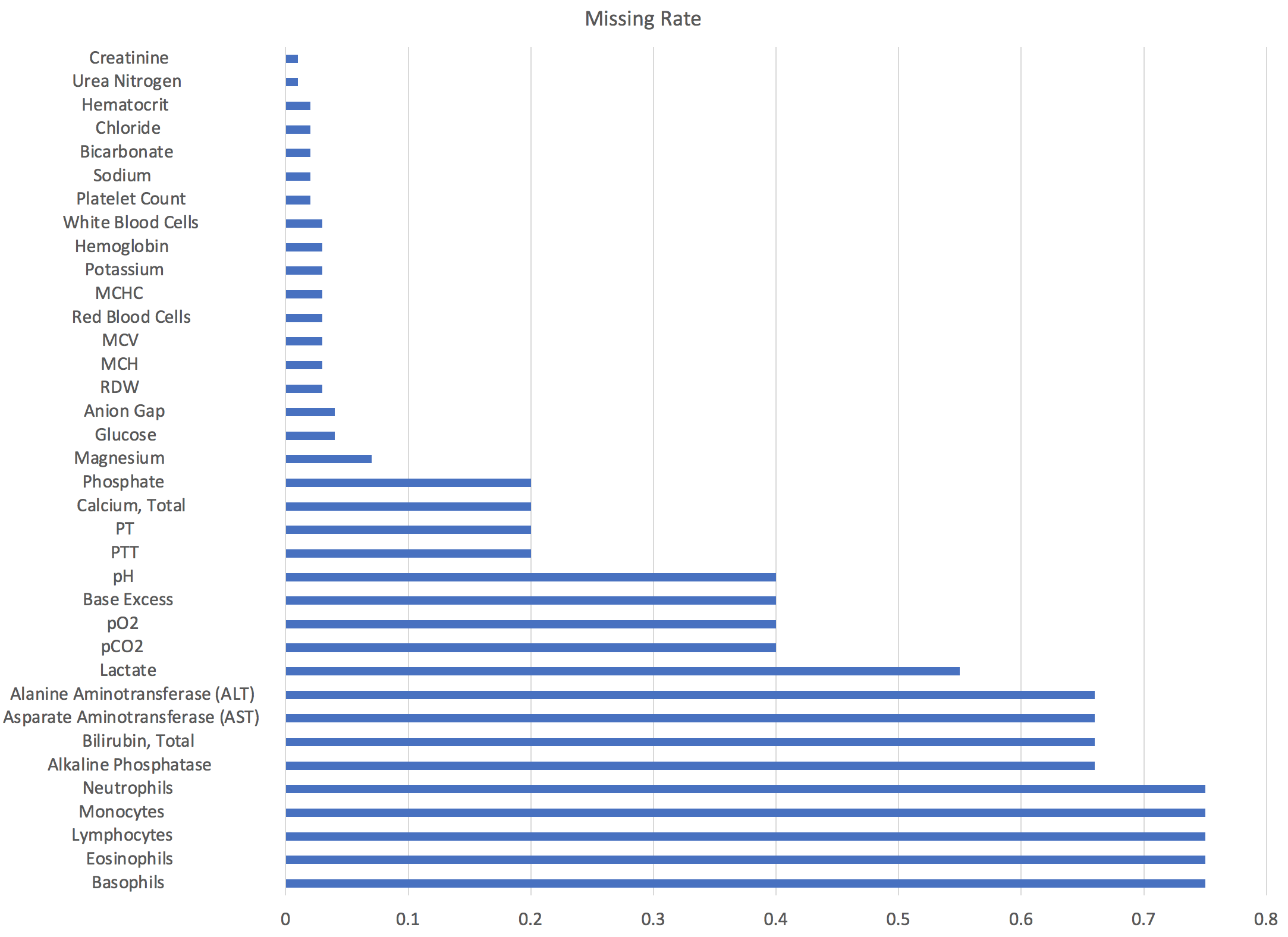}
\caption{ Missing Proportions of the 36 Common Laboratory Tests in MIMIC-III.}
\label{fig:Missing Proportions}
\end{figure}

\begin{table}
\caption{MIMIC-III Laboratory Data - Missing Imputation: Simulation of Blockwise Missing Data with $n = 5000$ and $p = 24$ under MAR. The missing rates of the features range from $10\%$ to $50\%$. Performance results are presented as mean $\pm$ standard deviation.}
\begin{tabular}{lccccc}
\hline
Missing Rate & 0.1 & 0.2 & 0.3 & 0.4 & 0.5 \\
\hline
Mean Imputation & $0.073 \pm 0.001$ & $0.073 \pm 0.001$ & $0.073 \pm 0.001$ & $0.073 \pm 0.001$ & $0.073 \pm 0.001$ \\
Multiple Imputation & $0.061 \pm 0.001$ & $0.063 \pm 0.001$ & $0.065 \pm 0.000$ & $0.067 \pm 0.001$ & $0.069 \pm 0.001$ \\
MI-KNN & $0.064 \pm 0.001$ & $0.066 \pm 0.001$ & $0.068 \pm 0.001$ & $0.070 \pm 0.001$ & $0.071 \pm 0.001$ \\
MissForest & $0.059 \pm 0.001$ & $0.061 \pm 0.001$ & $0.064 \pm 0.001$ & $0.066 \pm 0.001$ & $0.069 \pm 0.001$ \\
Soft & $0.080 \pm 0.002$ & $0.094 \pm 0.001$ & $0.098 \pm 0.006$ & $0.095 \pm 0.006$ & $0.092 \pm 0.004$ \\
Soft\_Mask & $0.097 \pm 0.003$ & $0.091 \pm 0.003$ & $0.097 \pm 0.004$ & $0.106 \pm 0.005$ & $0.125 \pm 0.007$ \\
GAIN & $0.061 \pm 0.001$ & $0.063 \pm 0.001$ & $0.065 \pm 0.001$ & $0.068 \pm 0.001$ & $0.070 \pm 0.001$ \\
GCMI & $\textbf{0.058} \pm \textbf{0.001}$ & $\textbf{0.061} \pm \textbf{0.001}$ & $\textbf{0.063} \pm \textbf{0.001}$ & $\textbf{0.066} \pm \textbf{0.001}$ & $\textbf{0.068} \pm \textbf{0.002}$ \\
\hline
\end{tabular}
\label{table: mimic}
\end{table}


\begin{table}
\caption{eICU Collaborative Research Database - Missing Imputation: Simulation of Blockwise Missing Data with $n = 5000$ and $p = 40$ under MAR. The missing rates of the features range from $10\%$ to $50\%$. Performance results are presented as mean $\pm$ standard deviation.}
\begin{tabular}{lccccc}
\hline
Missing Rate & 0.1 & 0.2 & 0.3 & 0.4 & 0.5 \\
\hline
Mean Imputation & $0.097 \pm 0.001$ & $0.097 \pm 0.001$ & $0.097 \pm 0.001$ & $0.097 \pm 0.001$ & $0.097 \pm 0.002$ \\
Multiple Imputation & $0.065 \pm 0.001$ & $0.070 \pm 0.001$ & $0.073 \pm 0.001$ & $0.078 \pm 0.001$ & $0.083 \pm 0.002$ \\
MI-KNN & $0.073 \pm 0.001$ & $0.076 \pm 0.001$ & $0.080 \pm 0.001$ & $0.084 \pm 0.001$ & $0.088 \pm 0.002$ \\
MissForest & $0.062 \pm 0.001$ & $0.067 \pm 0.001$ & $0.073 \pm 0.001$ & $0.078 \pm 0.001$ & $0.083 \pm 0.002$ \\
Soft & $0.072 \pm 0.001$ & $0.076 \pm 0.001$ & $0.079 \pm 0.001$ & $0.084 \pm 0.001$ & $0.090 \pm 0.002$ \\
Soft\_Mask & $0.100 \pm 0.001$ & $0.107 \pm 0.002$ & $0.112 \pm 0.002$ & $0.114 \pm 0.002$ & $0.113 \pm 0.002$ \\
GAIN & $0.076 \pm 0.001$ & $0.079 \pm 0.001$ & $0.082 \pm 0.001$ & $0.086 \pm 0.001$ & $0.112 \pm 0.002$ \\
GCMI & $\textbf{0.060} \pm \textbf{0.001}$ & $\textbf{0.065} \pm \textbf{0.002}$ & $\textbf{0.070} \pm \textbf{0.001}$ & $\textbf{0.076} \pm \textbf{0.001}$ & $\textbf{0.081} \pm \textbf{0.002}$ \\
\hline
\end{tabular}
\label{table: eicu}
\end{table}



\nocite{*}
\bibliography{sn-bibliography}

\end{document}